# Adaptive Artificial Intelligent Q&A Platform

Akram M.R[*, a)], Singhabahu C.P[*, b)], Saad M.S.M[*, c)], Deleepa P[*, d)], Nugaliyadde A[*, e)], Mallawarachchi Y[*, f)]

(* Department of Software Engineering)
Sri Lanka Institute of Information Technology,
Malabe 10115, Sri Lanka.
( [a)]rifhan.akram1, [b)]mails.chamara, [c)]saad.sahibjan, [d)]deleepa.p, [e)]anupiyanugel)@gmail.com
( [f)]yashas.m)@sliit.lk

*Abstract—* **The paper presents an approach to build a question and answer system that is capable of processing the information in a large dataset and allows the user to gain knowledge from this dataset by asking questions in natural language form. Key content of this research covers four dimensions which are; Corpus Preprocessing, Question Preprocessing, Deep Neural Network for Answer Extraction and Answer Generation. The system is capable of understanding the question, responds to the user's query in natural language form as well. The goal is to make the user feel as if they were interacting with a person than a machine.**

I. INTRODUCTION

The information growth rate in the world is increasing at a rapid rate. It has become impossible to keep up with the amount of information that is being added to any given domain. Due to the wide availability of digital input and output devices and the ease of use of these devices people are creating more and more raw data. If we are able to process this data into meaningful information fast, it would be possible for people to become more productive and to get the information they want faster.

This paper presents an approach that uses domain specific datasets to train deep learning models that are capable of answering domain specific queries. Deep learning can be defined a subset of machine learning techniques that uses non-linear information processing to identify and extract features and patterns in data, classification and transformations. There are three key reasons that deep learning has become so popular in the recent past. First, the hugely increased processing abilities and availability of general purpose GPUs, the vast amount of training data that has become available and the many advances made in the recent past in the field of deep learning that has made the task of training artificial neural networks more efficient [1].

Since deep neural networks are exemplary at recognising patterns and processing data extremely fast, we believe that by applying deep learning techniques to this problem domain we will be able to overcome many of the drawbacks of the other approaches. We will be able to have a higher accuracy because of the state of the art neural network training paradigms, reduce manual tasks by allowing an ANN to process and structure the corpus and automatically extract the required features and we will not need to use an underlying database engine so therefore we will not need to adopt or develop a different query language [2].

The initial step of this research is to pre-process domain related data that can be fed into a deep neural network for training and generates a machine learning model, further the inputs(questions) are being processed with word embedding techniques and output(answer) generation and formatting uses neural network techniques in order to present an output that is human friendly, in-depth explanation of the methodology is presented in the next section. The future works section discusses the suggested improvements that can be made for a robust and effective implementation.

II. BACKGROUND

An extensive prior study of various approaches used in the Q&A domain provided us with a clear roadmap for our research. This resulted in we choosing a deep learning based approach. Deep learning is a subset of techniques of machine learning.

*A. Literature Survey*

An extensive prior study of various approaches used in the Q&A domain provided us with a clear roadmap for our research. This resulted in we choosing a deep learning based approach. Deep learning is a subset of techniques of machine learning.

Machine learning is a field of Artificial Intelligence that has been gaining a lot of prominence in the current era. It is specifically to do with building systems that are able to learn by themselves without having to be programmed. Deep learning is a subset of techniques of machine learning. Deep learning allows multiple processing layers to breakdown the given data into smaller parts and learn the representations of these data [3]. Deep learning is the state of the art in areas such as speech recognition, natural language understanding, visual object recognition, etc. Convolutional neural networks have brought many breakthroughs in areas such as processing images, pictures and speech, whereas Recurrent networks have been extremely successful in areas such as processing text and speech.

QA is a well researched area from the point of NLP (Natural Language Processing) research. QA has mostly been used to develop intricate dialogue systems such as chatbots and other systems that mimic human interaction [4]. Traditionally most of these systems use the tried methods of parsing, part-of-speech tagging, etc that come from the domain of NLP research. While there is absolutely nothing wrong with these techniques, they do have their limitations. [5] W.A. Woods et al. shows how we can use NLP as a front end for extracting information from a given query and then translate that into a logical query which can then then be converted into a database query language that can be passed into the underlying database management system. In addition to that there needs to be a lexicon that functions as an admissible vocabulary of the knowledge base so that it is possible to filter out unnecessary terminology. The knowledge base is processed to an ontology that breaks it down into classes, relations and functions [6]. Natural Language Database Interfaces (NLDBIS) are database systems that allow users to access stored data using natural language requests. Some popular commercial systems are IBM's LanguageAccess and Q&A from Symantec [7].

Information retrieval (IR) is another technique that has been used to address the problem of QA. With IR systems pay attention to the organisation, representation and storage of information artifacts such that when a user makes a query the system is able to to return a document or a collection of artifacts that relate to the query [8]. Recent advances in OCR and other text scanning techniques have meant that it is possible to retrieve passages of text rather than entire documents. However IR is still widely seen as from the document retrieval domain rather than from the QA domain.

Template based question answering is another technique that has been used for QA and is currently being used by the START system which has answered over a million questions since 1993 [9]. START uses natural language annotations to match questions to candidate answers. An annotation will have the structure of 'subject-relationship-object' and when a user asks a question, the question will be matched to all the available annotation entries at the word level (using synonyms, IS-A, etc) and the structure level. When a successful match is found, the annotation will point to an information segment which will be returned as the answer. When new information resources are incorporated into the SMART system, the natural language annotations have to be composed manually [10]. START uses Omnibase as the underlying database system to store information and when the annotation match is found, the database query must be used to retrieve the information. While this system has been relatively successful, it requires a lot of preprocessing which must be done manually.

Our literature survey has found that the QA domain has an active community of researchers and many different approaches have been tried to tackle this problem. While the problem of QA is a very old one, the origins of the problem can be traced back as far as the 1960's, using our access to cheaper and better computational power and newer techniques in data processing we believe we can attempt to solve this issue using a different set of tactics.

III. METHODOLOGY

To further narrow down our scope we have chosen to build a question answer system using the Cornell movie quotes corpus and not addressing open domain questions. The scope has been thus narrowed to, first, increase the accuracy of answers provided and second, to ensure that the project can be completed in the allocated time.

In order to achieve the main objective, we have broken it down into four different sub objectives. Each of these sub objectives form a critical part of the system and carry out a critical function. They also have deeply integrated deep learning techniques in each component, which we have described in great detail in the methodology section. In the next section we have a brief overview of what each of the sub objectives are supposed to accomplish. Figure 1.0 shows how the components have been arranged and how they will interact with each other.

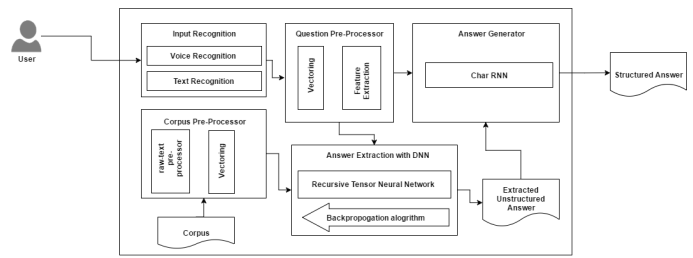

Figure 1.0 System Architecture

*A. Corpus preprocessing*

One of the first things we need to do with the data corpus is to pre-process it, so that we can use a training network to create the model. Initially the corpus will be unstructured text data, which can be understood by humans, not by machines. Word Embedding [11] is also used for semantic parsing, to extract the meaning from text to enable Natural Language Understanding. For a language model to understand the meaning of a word, it need to know the contextual similarity of words. For example if we tend to find diseases in sentences, where diabetes, diarrhea, HIV should be of close proximity.

Batching is the process of breaking down the training data set into multiple batches and adjusting the weights after each batch has been used to train the network. For example, if a training set has 1050 data points, we can train the neural network with 100 batches at a time. This increases the efficiency of the training process [12], as opposed to the online training process. We chose to use this method because of the limitation of hardware resources and time constraints.

An epoch is reached during the training process when the entire training dataset has been used to train the neural network and then the process is repeated with the same training dataset. If this is done too much, the model will end up overfitting the training dataset. So we chose to train our model over 30 epochs.

*B. Question pre-processing*

The question that the user is asking need to be preprocessed as well along with the other text data. Word embedding maps each word to a vector space.The Embedding layer will map each token from the question to its corresponding vector space, which preserves the contextual similarity of words in the vector space. The embedding layer in the question processing component can be done through a popular pretrained word embedding model known as word2vec or glove (exact model will be chosen based on further trial and error) [13].

Since we are using the sequence-to-sequence model for training, we need to encode the question sequence into a fixed-size state vector. When creating the LSTM encoder/decoder cells, we use the Tensorflow framework tf.contrib.rnn.BasicLSTMCell. The encoding process is handled by the Tensorflow framework. A deeper explanation of this will be given in the next section.

*C. Deep Neural Network for Answer Extraction*

This component basically acts as the part where the actual decision making happens. The primary objective of this module is to generate an answer to the pre-processed question by using the structured data available in the pre-processed corpus. In this research we use a recurrent neural network (RNN) to extract the answers from the corpus. In previous researches it was found that RNN's perform much better than other neural networks when it comes to text processing.

A sequence to sequence model consists of two recurrent neural networks. One is an encoder that will deal with the input and the other is a decoder that will generate the output. Figure 2.0 shows the general architecture for such a network. As shown, generally the decoder cell will start after a custom <go> tag.

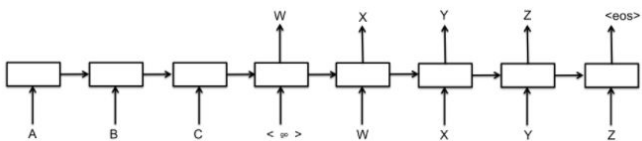

Figure 2.0 An RNN cell can be of type GRU (Gated Recurrent Unit) or LSTM (Long Short Term Memory). [14]

In our DNN implementation, we are using LSTM's because they are better at dealing with vanishing gradients and have some memory capacity [15]. We are using two LSTM layers with a hidden size of 256. This was deemed the best approach after reading through research papers. We did not have the resources or time to test with different configurations.

In the system this model gets its inputs from the question pre-processing component and outputs the generated tensor into the answer generation component in order to generate a human readable answer.

*D. Answer Generation*

This component is required because the corpus will not always have the answer in the required format.The challenge with generating natural language from a single piece of text is that it is the reverse approach to what is taken by most other researchers. Usually deep neural networks are used to identify patterns and are used as predictive models. In this case the DNN must be used as a generative model instead.

After the question is encoded and fed into the LSTM, the neural network will process the output and give the answer that it located via the decoder. However we need to map the decoder output into a word vector for better representation. For this we use a Tensorflow function: tf.contrib.legacy_seq2seq.embedding_rnn_seq2seq. This function returns the decoder outputs as an embedded model [16]. This is gives us a set of words that the neural network thinks is the answer for the question. One of the future tasks we have is to better process this output.

IV. RESULTS AND DISCUSSION

The intended use of this system is to allow the user to ask related questions based on the defined domain (i.e on the plugged dataset for training. eg: medical emergency, cornel, ubuntu) in natural language form and the platform will generate the most accurate answer and provide that answer in natural language form as well. The idea is to simulate a situation where the user is interacting with a person in the medical profession as close as possible.

The system was tested with multiple trained datasets such as Cornell movie dialogs [17], Ubuntu Q&A forum [18] and OpenSubs [19].

It is nearly impossible to make an algorithm to understand a language and respond to questions. To overcome these type of issues artificial neural networks were introduced. ANN is an information processing paradigm that is inspired by the way biological nervous systems. One of the famous ANN technique named recurrent neural networks was used. One of the appeals of RNNs is the idea that they might be able to connect previous information to the present task, such as using previous video frames might inform the understanding of the present frame. But with RNN this capability is limited. Predicting the last word in "the clouds are in the sky" is possible since it does not require any additional contextual understanding. But for "I grew up in France… I speak fluent French." It requires more memory and understanding of context. This is the reason to use Long Short Term Memory networks, called as "LSTMs" – are a special kind of RNN, capable of learning long-term dependencies. With the use of these techniques results were given as in Figure 3.0.

```
Q: Hi
A: Hi.

Q: What is your name ?
A: Laura.

Q: What does that mean ?
A: I dunno.

Q: How old are you ?
A: thirty-five.

Q: Will Google hire me ?
A: No.
```

Figure 3.0

But there were some edge cases where the system failed to produce results as expected as in Figure 4.0. This happens due to training longer sentences or conversations. In that case, the network is much slower to train. After 250000 iterations with maxLength=10, the network still mainly gives evasive answers (by answering most of the open ended questions by 'I don't know') but it seems to have integrated some basic semantic.

```
Q: Who is Laura ?
A: My brother.

Q: Say 'goodbye'
A: Alright.

Q: What is cooking ?
A: A channel.

Q: Can you say no ?
A: No.

Q: Two plus two
A: Manny...
```

Figure 4.0

V. RESEARCH FINDING

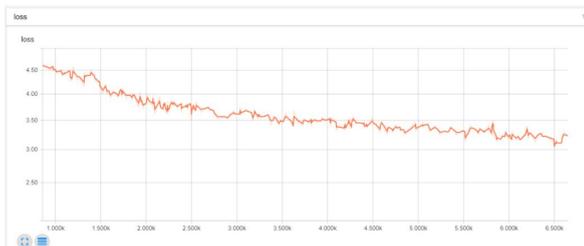

Figure 5.0 - Loss Function Graph from Tensor Board

Loss functions presents the information about training against testing. The graph in Figure 5.0 it shows the loss reduction as the system trains for 30 epochs. An epoch is one complete presentation of the data set to be learned by a learning machine. To get better accuracy the model needs to be trained until the loss graph reaches it lowest level. But it should not increase again since it will introduce an issue called overfitting as represented in Figure 6.0 by the red line. In overfitting, a statistical model describes random error or noise instead of the underlying relationship. That means it will bound to the given training dataset and making any questions out of the dataset might provide inaccurate answers.

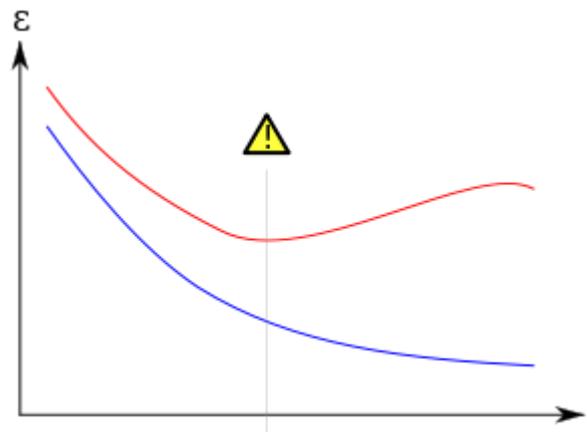

Figure 6.0 - Representation of Overfitting [20]

The purpose of word embedding [21] is to extract meaning from text to enable natural language understanding. To increase the semantic meaning of contextually similar of words, the corpus can be preprocessed by merging multi-word terms. Such as the terms Human Immunodeficiency Virus transformed to human_immunodeficiency_virus to understand it as a single term than three separate terms during the training process to enable improved semantic parsing.

VI. FUTURE WORK

As improvements in the future that can enable the proposed approach to perform better and cater to additional features, we propose two main improvements.

The current implementation can only answer queries for a single domain utilizing a single trained machine learning model. This limitation is because the system is not designed to handle multiple machine learning models in a single running instance of the system. We propose to perform classification at question level where the input (question) will be classified to understand its type/category and according to its relevant category the question will be passed over to its relevant domain specific data model for processing. The classes for classification are directly the domains supported by the running instance of the system. Due to the pluggable architecture of the system the implementation of the above improvement can be made.

The other suggested improvement is to support dependencies between questions. Currently questions are independent from each other. To enable dependencies between questions a straightforward approach would be to feed all previous questions and answers to the encoder before generating the answer. Further, the network should be retrained on QA pairs which make up a dialogue instead of just individual QA.

## VII. Conclusion

Overall from this research we have been able to gain an understanding into the complexities of Deep Neural Networks and have managed to create an application that integrates with a DNN to allow a user to interact with a knowledge base in QA format. The main challenges we faced was to understand the concepts and the time that it takes to build a DNN and to train it. Here it was important to understand the importance of not overtraining the model by keeping an eye on the loss function. The accuracy of the output was another challenge and it requires a lot more work to understand how we can improve the accuracy of the neural network. Due to time constraints, it is not possible for us to divert from the current approach and to adopt a new one. We have identified some improvements for future work that will help us create a more viable product from this application.

## References

[1] L. Deng and D. Yu, "Deep Learning: Methods and Applications", Foundations and Trends in Signal Processing, vol. 7, no. 34, pp. 197–387, 2013.

[2] S. Wang, Interdisciplinary computing in Java programming language, 1st ed. Boston, Mass.: Kluwer Academic, 2003.

[3] Y. LeCun, Y. Bengio and G. Hinton, "Deep learning", Nature, vol. 521, pp. 436 – 444, 2015.

[4] Silvia Quarteroni, "A Chatbot-based Interactive Question Answering System", 11th Workshop on the Semantics and Pragmatics of Dialogue, 2007.

[5] W.A Woods, R.M. Kaplan and B. Nash-Webber, "The lunar sciences natural language information system", BBN Rep. 2378, Bolt Beranek and Newman, Cambridge, Mass., USA, 1977.

[6] T.R. Gruber, "A translation approach to portable ontology specifications", Knowledge Acquisition, 5 (2), 1993.

[7] R. Dale, H. Moisl and H. Sommers, Handbook of Natural Language Processing, 1st ed. New York: Marcel Dekker AG, 2006, pp. 215 - 250.

[8] L. Hirschman and R. Gaizauskas, "Natural language question answering: the view from here", Natural Language Engineering, 7 (4), 2001, pp. 275-300.

[9] "The START Natural Language Question Answering System", Start.csail.mit.edu, 2017. [Online]. Available: http://start.csail.mit.edu/index.php. [Accessed: 26- Mar- 2017].

[10] B. Katz, G. Borchardt and S. Felshin, "Natural Language Annotations for Question Answering", Proceedings of the 19th International FLAIRS Conference (FLAIRS 2006), 2006.

[11] A. Colyer, "The amazing power of word vectors", the morning paper, 2016.[Online].Available:https://blog.acolyer.org/2016/04/21/the-amazing-power-of-word-vectors/.

[12] S. Ioffe and C. Szegedy, "Batch Normalization: Accelerating Deep Network Training by Reducing Internal Covariate Shift", 2017.

[13] Ruder, Sebastian. "On Word Embeddings - Part 3: The Secret Ingredients Of Word2vec". Sebastian Ruder. N.p., 2017. Web. 26 Mar. 2017.

[14] "Sequence-to-Sequence Models", TensorFlow, 2017. [Online]. Available: https://www.tensorflow.org/tutorials/seq2seq. [Accessed: 05- Aug- 2017].

[15] "Recurrent Neural Network Tutorial, Part 4 – Implementing a GRU/LSTM RNN with Python and Theano", WildML, 2017. [Online]. Available: http://www.wildml.com/2015/10/recurrent-neural-network-tutorial-part-4-implementing-a-grulstm-rnn-with-python-and-theano/. [Accessed: 08- Aug- 2017].

[16] "tf.contrib.legacy_seq2seq.embedding_rnn_seq2seq", TensorFlow, 2017. [Online]. Available: https://www.tensorflow.org/api_docs/python/tf/contrib/legacy_seq2seq/embedding_rnn_seq2seq. [Accessed: 11- Aug- 2017].

[17] "Cornell Movie-Dialogs Corpus", Cs.cornell.edu, 2017. [Online]. Available: http://www.cs.cornell.edu/~cristian/Cornell_Movie-Dialogs_Corpus.html. [Accessed: 05- Aug- 2017].

[18] "The Ubuntu Dialogue Corpus", Dataset.cs.mcgill.ca, 2017. [Online]. Available: http://dataset.cs.mcgill.ca/ubuntu-corpus-1.0/. [Accessed: 05- Aug- 2017].

[19] "OpenSubtitles", Opus.lingfil.uu.se, 2017. [Online]. Available: http://opus.lingfil.uu.se/OpenSubtitles.php. [Accessed: 05- Aug- 2017].

[20] "Overfitting", En.wikipedia.org, 2017. [Online]. Available: https://en.wikipedia.org/wiki/Overfitting. [Accessed: 08- Aug- 2017].

[21] "An overview of word embeddings and their connection to distributional semantic models - AYLIEN", AYLIEN. [Online]. Available: http://blog.aylien.com/overview-word-embeddings-history-word2vec-cbow-glove/.